\documentclass[10pt,twocolumn,letterpaper]{article}

\usepackage{wacv}              

\definecolor{wacvblue}{rgb}{0.21,0.49,0.74}
\usepackage[pagebackref,breaklinks,colorlinks,allcolors=wacvblue]{hyperref}

\usepackage[normalem]{ulem}
\usepackage{multirow}
\usepackage{subcaption}
\usepackage{enumitem}
\usepackage{graphicx}
\usepackage{algorithm}
\usepackage{algpseudocode}
\algtext*{EndIf}
\algtext*{EndFor}
\algtext*{EndFunction}
\algtext*{EndWhile}
\usepackage{tikz}

\tikzset{algpxIndentLine/.style={draw=blue,dashed}}

\usepackage{xspace}
\usepackage{xcolor,colortbl}
\usepackage{wrapfig}
\usepackage{comment}
\usepackage{bbding}
\newcommand{\colorcheck}{\textcolor{green!70!black}{\CheckmarkBold}}
\newcommand{\grayrow}{\rowcolor[gray]{.9}}
\definecolor{baselinecolor}{gray}{.95}

\algrenewcommand{\algorithmiccomment}[1]{\hfill\textcolor{gray}{\(\triangleright\) #1}}

\definecolor{best}{HTML}{E4EAFF}
\definecolor{second}{HTML}{FFE8D9}
\definecolor{white}{HTML}{FFFFFF}
\definecolor{ours}{HTML}{56ab15}

\newcommand{\best}[1]{\colorbox{best}{#1}}
\newcommand{\second}[1]{\colorbox{second}{#1}}
\newcommand{\white}[1]{\colorbox{white}{#1}}
\newcommand{\ours}[1]{\textcolor{ours}{#1}}

\usepackage{wrapfig}
\usepackage{booktabs}

\def\wacvPaperID{2061} 
\def\confName{WACV}
\def\confYear{2026}

\title{Unsupervised Discovery of Long-Term Spatiotemporal Periodic Workflows in Human Activities}

\author{
Fan Yang\textsuperscript{1}, 
Quanting Xie\textsuperscript{3}, 
Atsunori Moteki\textsuperscript{2}, 
Shoichi Masui\textsuperscript{2}, \\ 
Shan Jiang\textsuperscript{2},
Kanji Uchino\textsuperscript{1},
Yonatan Bisk\textsuperscript{3},  
Graham Neubig\textsuperscript{3}\\
\normalsize
\textsuperscript{1} Fujitsu Research of America, USA  
\textsuperscript{2} Fujitsu Limited, Japan 
\textsuperscript{3} Carnegie Mellon University, USA
\\
}

\begin{document}
\twocolumn[{%
\renewcommand\twocolumn[1][]{#1}%
\maketitle
\begin{center}
    \centering
    \captionsetup{type=figure}
    \includegraphics[width=\textwidth,height=5cm]{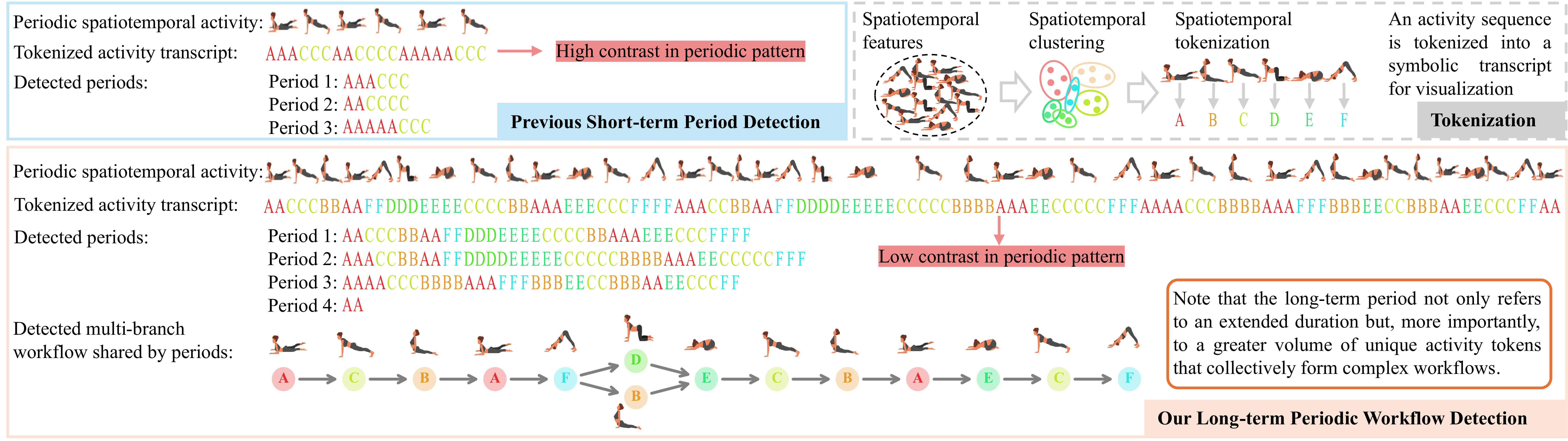}
    \captionof{figure}{\textbf{Periodic spatiotemporal activity in a yoga example}. While existing studies focus on short-term periods with simple structures and high-contrast patterns, we investigate a novel direction on long-term periods that involve complex workflows with low-contrast patterns.}
    \label{fig:overview}
\end{center}%
}]

\begin{abstract}
Periodic human activities with implicit workflows are common in manufacturing, sports, and daily life. While short-term periodic activities---characterized by simple structures and high-contrast patterns---have been widely studied, long-term periodic workflows with low-contrast patterns remain largely underexplored. 
To bridge this gap, we introduce the first benchmark comprising 580 multimodal human activity sequences featuring long-term periodic workflows. 
The benchmark supports three evaluation tasks aligned with real-world applications: unsupervised periodic workflow detection, task completion tracking, and procedural anomaly detection.
We also propose a lightweight, training-free baseline for modeling diverse periodic workflow patterns.
Experiments show that: (i) our benchmark presents significant challenges to both unsupervised periodic detection methods and zero-shot approaches based on powerful large language models (LLMs); (ii) our baseline outperforms competing methods by a substantial margin in all evaluation tasks; and (iii) in real-world applications, our baseline demonstrates deployment advantages on par with traditional supervised workflow detection approaches, eliminating the need for annotation and retraining.
Our project page is \url{https://sites.google.com/view/periodicworkflow}.
\end{abstract}

    
\vspace{-0.5cm}
\section{Introduction}
\label{sec:intro}

The rhythmic repetition of activities, characterized by periodic spatiotemporal workflows, is ubiquitous in many aspects of human life and work, ranging from the structured routines of a factory production line to the cyclical movements of athletes in training~\cite{lorenz2010periodization, bompa2019periodization, bordel2022recognizing, cheng2023periodic, li2010mining, ghosh2017finding, song2022robust}. Detecting and understanding these periodic workflows offers immense potential for optimizing working efficiency, enhancing training performance, improving motion synthesis, and facilitating knowledge transfer to new learners or robots~\cite{lee2006human, lin2018classification, abedi2023rehabilitation, yang2018learning, pertsch2025fast}.

\begin{figure*}[t!]
  \centering
  \includegraphics[width=\linewidth]{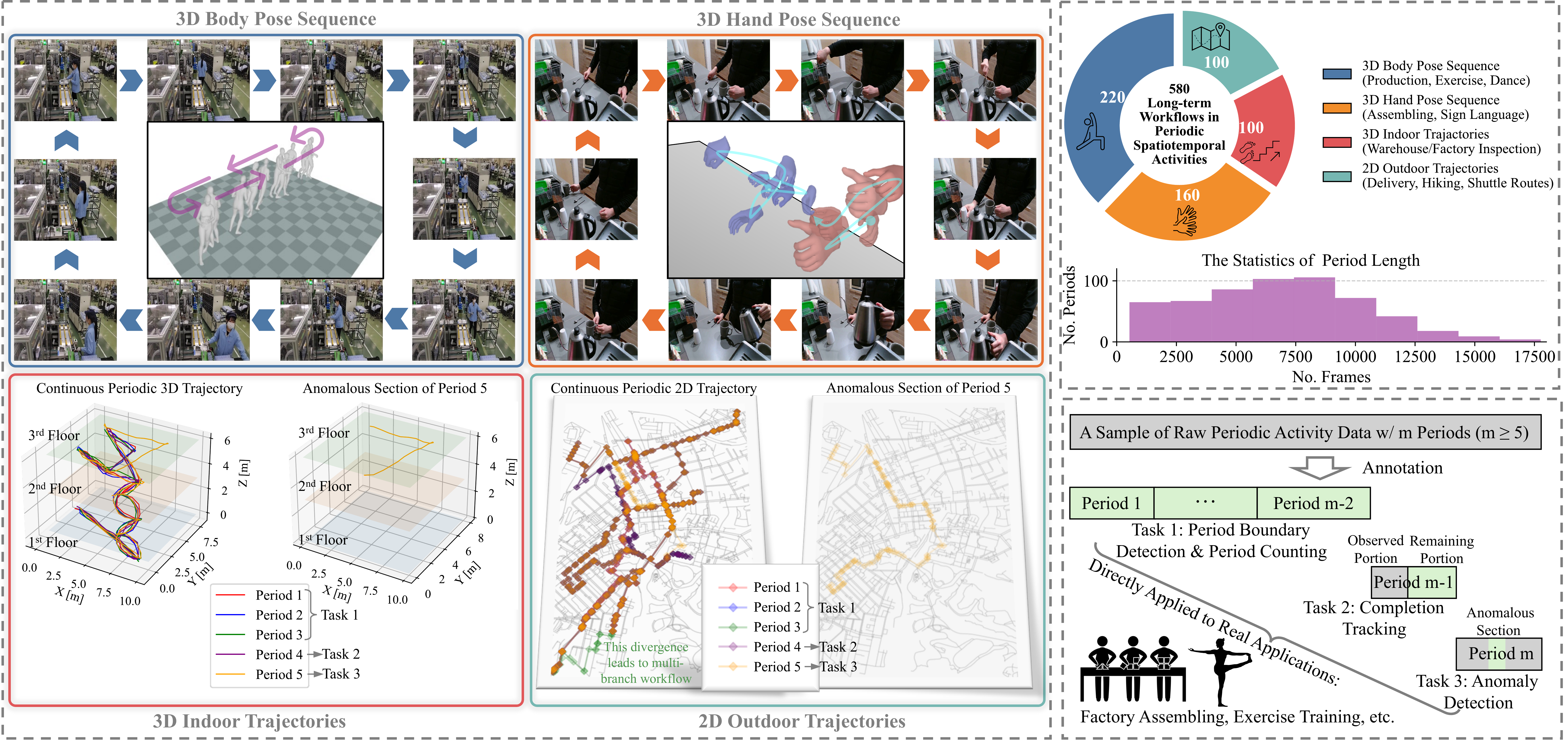}
  \vspace{-0.5cm}
  \caption{\textbf{Left:} Illustration of our benchmark. \textbf{TopRight:} Statistics of our dataset. \textbf{BottomRight:} Annotation of our benchmark. We collected 580 long-term workflows of periodic human activities, \textbf{characterized by a compact dataset yet encompassing a wide variety of real-world periodic tasks}, including factory production, exercise training, and shuttle routes. }
  \label{fig:benchmark}
 \vspace{-0.2cm}
\end{figure*}

Current methods for periodic activity detection generally focus on short-term periods with clear start and end points, often relying on supervised learning to detect high-contrast patterns between periods~\cite{panagiotakis2018unsupervised, dwibedi2020counting, zhang2021repetitive, hu2022transrac, yao2023poserac, yang2023simper, destro2024cyclecl, li2024repetitive, li2024efficient, sinha2024every, lu2024fca, luo2024rethink, dwibedi2024ovr, li2025repetitive}. However, many real-world activity periods are long-term, consisting of ordered activity tokens that form workflows, yet remain largely unexplored. For instance, in periodic workflows, identical action tokens may appear multiple times at different positions within a period, which complicates period boundary identification---an issue often overlooked in prior studies (Fig.~\ref{fig:overview}, Sec.~\ref{sec:related}). Additionally, existing studies on human activity segmentation typically divide activities into multiple tokens based on a workflow. However, these studies often ignore the period detection, assuming a predefined period for segmentation~\cite{lea2017temporal, richard2018nnviterbi, farha2019ms, yang2019make, MuCon2021, li2021temporal, asformer2021, ding2022leveraging, rahaman2022generalized, behrmann2022unified, du2022fast, su2024TAS, xu2024efficient, spurio2024hierarchical}.

To bridge this gap, we introduce a pioneering benchmark for unsupervised detection of long-term periodic spatiotemporal workflows in human activities. Specifically, we collected $580$ human activity sequences, \textbf{with a small size but a rich diversity to cover most real-world periodic activities}, such as factory production, exercise training, and shuttle routes. Our dataset covers rich modalities, including 3D body/hand pose sequences, 3D indoor trajectories, and 2D outdoor trajectories (Fig.~\ref{fig:benchmark}). With our provided spatiotemporal features, we have designed three evaluation tasks. 
In the first task, we introduce an activity sequence consisting of more than three periods, requiring unsupervised identification of period boundaries and counting.
The second task focuses on completion tracking, which involves using the extracted workflow to estimate the remaining portion of a new period when only part of that period is available. The final task utilizes the extracted workflow to localize procedural anomalies within a new period. These tasks correspond to real-world applications, such as assisting production line workers in optimizing schedules and reducing errors.

We provide a novel baseline method for our benchmark, starting with a spatiotemporal tokenization to tokenize all frames of an activity into a symbolic transcript using both hard and soft tokens. The hard token for each frame is determined by the symbol of its nearest token feature centroid, while the soft token consists of a vector of normalized distances to all token feature centroids. We then form a 2D matrix from the soft tokens of all frames and analyze it in the frequency domain. We identify a rough periodic scale along the temporal axis and apply it to segment the entire transcript. Finally, we propose an innovative sequential mining algorithm to distill the common pattern within all segmented hard-tokenized transcripts. This approach yields unified periodic activity workflows with precisely defined boundaries.
\textbf{Our method is fully unsupervised, eliminating the need for textual activity labels, which are often difficult to define for complex human activities}~\cite{li2021towards}. For example, whereas traditional action-to-text tokenization requires manual assignment of esoteric labels to actions (\eg, `pirouette' in ballet), our scheme uses arbitrary symbols (\eg, `A'$\sim$`Z'). This abstraction supports all downstream applications in our benchmark, without the need to explicitly know the activity names.

We summarize our contributions in three main areas:
\begin{itemize}[nosep, leftmargin=*]
  \item Benchmark Innovation: We introduce the first benchmark featuring long-term periodic spatiotemporal workflows of human activity, gathered from diverse real-world settings. This benchmark effectively mirrors the complexities encountered in practical applications and opens exciting research avenues in the field of human activity studies.
  \item Algorithm Development: We developed an innovative baseline to tackle challenges posed by our benchmark. Our evaluation demonstrates its promising performance, establishing a robust methodology for future research endeavors and real-world applications.
  \item Real-World Application: We introduce three evaluation tasks to address real-world applications, including unsupervised period detection, task completion tracking, and procedural anomaly detection. We illustrate the practical value of our work through a successful deployment in a real-world factory production line. 
\end{itemize}

\section{Related Works}
\label{sec:related}

\subsection{Short-term Periodic Activity Detection (S-PAD)}
\label{sec:short_term}
Existing research on periodic human activity detection primarily focuses on short-term repetitive actions~\cite{panagiotakis2018unsupervised, dwibedi2020counting, zhang2021repetitive, hu2022transrac, yao2023poserac, yang2023simper, destro2024cyclecl, li2024repetitive, li2024efficient, sinha2024every, lu2024fca, luo2024rethink, dwibedi2024ovr, li2025repetitive}. 
Methodologically, a common approach across these studies involves employing neural networks to encode spatiotemporal features and constructing a temporal self-similarity matrix (TSM) for periodicity detection.  Furthermore, some investigations~\cite{hu2022transrac, yao2023poserac, li2024repetitive, li2025repetitive} utilize transformers to capture temporal dependencies through supervised learning, while others~\cite{panagiotakis2018unsupervised, yang2023simper, destro2024cyclecl} concentrate on learning periodic patterns without manual annotations or prior action knowledge.  
Beyond unimodal approaches, another line of research~\cite{zhang2021repetitive} integrates visual and auditory cues for multimodal period detection.
In terms of datasets, studies such as \cite{zhang2021repetitive, hu2022transrac, yao2023poserac, yang2023simper, sinha2024every} have introduced new datasets to advance research in this field. While these methods and datasets have proven effective for analyzing short repetitive actions, they have not addressed the complexities of long-term periodic activities characterized by embedded workflows. Our work bridges this critical gap by introducing both a novel benchmark and a robust baseline method specifically tailored for such challenging scenarios (Tab.~\ref{tab:benchmark_comparison}).

\subsection{Single-period Activity Segmentation (SAS)}
\label{sec:no_period}
Research in human activity segmentation aims to partition a single-period sequence of activities into distinct, bounded activity tokens that conform to specific workflows. We compare our work with prior SAS studies~\cite{fathi2011learning, stein2013combining, kuehne2014language, gao2014jhu, li2015delving, alayrac2016unsupervised, tang2019coin, sener2022assembly101} in Tabs.~\ref{tab:benchmark_comparison} and \ref{tab:SAS_comparison}.
Traditional SAS methods generally rely on labeled data to align activity features and tokens with semantic tags. While fully-supervised SAS~\cite{lea2017temporal, farha2019ms, asformer2021} requires exhaustive annotations of boundaries and labels for all activity tokens, semi-supervised SAS~\cite{richard2018nnviterbi, MuCon2021, li2021temporal, ding2022leveraging, rahaman2022generalized, behrmann2022unified, yang2023weakly,su2024TAS, xu2024efficient} reduces annotation costs, as it requires only a limited number of annotated timestamps during training. Nevertheless, both approaches face a critical limitation: the need to annotate new activity semantic tags and retrain the model when applied to novel scenarios. This requirement poses a significant barrier to scalability, hindering the generalization of these methods across diverse applications. 

Unsupervised SAS~\cite{sener2018unsupervised, kukleva2019unsupervised, sarfraz2021temporally, kumar2022unsupervised, du2022fast,xu2024temporally, li2024otas, spurio2024hierarchical} overcomes this limitation by utilizing clustering-based features to represent action tokens, thereby circumventing the necessity for explicit mappings between activity features and textual semantic tags. Even without semantic tags, this approach can efficiently extract segmentation patterns (\ie, workflows) from a single-period activity sequence.
However, unsupervised SAS often assumes a single period is available for segmentation. This limits its applicability in scenarios where periodicity should be inferred. Our approach bridges this gap by jointly integrating period detection and unsupervised workflow detection. 

\begin{table}[t]
	\footnotesize
	\setlength{\tabcolsep}{1.8pt}
	\centering
	\begin{tabular}{l|ccc}
	\toprule
	\textbf{Benchmarks}& \textbf{w/ Periods} & \textbf{w/ Workflow}  & \textbf{Learning}  \\
	\midrule
	S-PAD (Sec.~\ref{sec:related}) &  \CheckmarkBold & \XSolidBrush & Supervised  \\
	SAS  (Sec.~\ref{sec:related}) & \XSolidBrush & \CheckmarkBold   & Supervised / Unsupervised \\
	\grayrow
	\textbf{Ours} (Sec.~\ref{sec:dataset_and_metrics}) & \colorcheck & \colorcheck & Unsupervised \\
	\bottomrule
	\end{tabular}
    \vspace{-0.1cm}
	\caption{\textbf{Comparison of S-PAD, SAS, and our benchmarks.}}
	\label{tab:benchmark_comparison}
	\vspace{-0.2cm}
\end{table}

\begin{table}[t]
	\footnotesize
	\setlength{\tabcolsep}{1.8pt}
	\centering
	\begin{tabular}{l|cccc}
	\toprule
	\multirow{2}{*}{\textbf{Benchmark}} & \multirow{2}{*}{\shortstack[c]{\textbf{No. multi-branch} \\ \textbf{workflows}}} & \multicolumn{3}{c}{\textbf{Unsupervised Tasks}} \\
\cline{3-5}
& & \textbf{Period} & \textbf{Completion} & \textbf{Anomaly} \\
	\midrule
	GTEA~\cite{fathi2011learning} & 7 & \XSolidBrush & -  & - \\
    Breakfast~\cite{kuehne2014language} & 10 & \XSolidBrush & -  & - \\
    50Salads~\cite{stein2013combining} & $<$50 & \XSolidBrush & -  & - \\
    Assembly101~\cite{sener2022assembly101} & 362 & \XSolidBrush & -  & \CheckmarkBold \\
	\grayrow
	\textbf{Ours} & 580 & \colorcheck & \colorcheck & \colorcheck \\
	\bottomrule
	\end{tabular}
    \vspace{-0.1cm}
	\caption{\textbf{Comparison of workflow analysis with SAS studies.}}
    \vspace{-0.5cm}
	\label{tab:SAS_comparison}
\end{table}

\subsection{Periodic Spatiotemporal Data Mining}
\label{sec:period_data_mining}
Prior research in periodic spatiotemporal data mining, exemplified by studies such as \cite{cao2007discovery, jindal2013spatiotemporal, zhang2015periodic, zhang2018periodic, zeng2024mining}, has predominantly focused on analyzing 2D spatiotemporal GPS trajectories, primarily within traffic contexts. These investigations typically examine sequences of locations visited at regular intervals over time. However, our work diverges significantly from this established paradigm by encompassing a wider spectrum of human activities, spanning diverse domains such as factory production line operations and exercise training regimens.  Our dataset also offers greater modality diversity, incorporating 3D human body and hand pose sequences, along with 3D indoor and 2D outdoor trajectory data.
Furthermore, conventional approaches frequently utilize trajectories restricted to fixed temporal boundaries, such as 24-hour cycles. In contrast, our dataset captures periodic human activities with flexible temporal boundaries. 
Moreover, many existing methodologies inadvertently mirror the principles of process mining~\cite{van2012process, fournier2021finding}, where identical activity tokens positioned at different locations within a workflow can lead to self-loops in the graph, potentially losing long-term temporal dependencies in procedural analysis.
Importantly, this field currently lacks a publicly accessible unified dataset and evaluation metrics. To address this gap, we will release a comprehensive benchmark to promote  further research in the field.

\section{Our Benchmark}
\label{sec:dataset_and_metrics}
\noindent\textbf{Dataset.} As Fig.~\ref{fig:benchmark} shows, we collected 580 long-term workflows of periodic human activities by identifying real-world scenarios where such patterns naturally occur. For instance, while prior SAS datasets focus on single instances (\eg, making one cup of coffee), our dataset captures periodic workflows (\eg, repeatedly making multiple cups of coffee).
Each activity sequence in our dataset is annotated with the number of periods, period boundaries,  a unique workflow,  and corresponding activity tokens (Note: the use of arbitrary tokens does not affect evaluation under our benchmark setting). To ensure sufficient temporal coverage, each sequence contains at least five and at most eight periods. For task 2 (completion tracking), we randomly truncate a portion of the second-last period; for task 3 (anomaly detection), we inject an anomaly into the final period. This design strikes a balance between complexity and tractability, allowing the study of realistic periodic workflows while keeping evaluation computationally manageable.
To mitigate bias from various activity feature extraction models while protecting privacy, we provide processed spatiotemporal features---primarily motion trajectories---for periodic workflow analysis. These features are synthesized to reflect natural patterns of human activities and corresponding workflow structures, and they include accurate ground-truth labels. For practical applications, it is recommended to use off-the-shelf human body and hand pose estimators~\cite{shin2024wham, shen2024world, wang2025tram, potamias2024wilor, zhang2025hawor} to obtain 3D body and hand sequences in small-scale environments. Indoor human 3D trajectories can be captured using Radio Frequency Identification (RFID)~\cite{ni2003landmarc, zafari2019survey} and visual-inertial odometry (VIO)~\cite{AppleARKit}, while outdoor 2D trajectories are recorded via Global Navigation Satellite System (GNSS).

\noindent\textbf{Evaluation Tasks.}  We define three evaluation tasks using the provided spatiotemporal features: \textbf{1.} detect period counts and boundaries on normal periods through unsupervised methods; \textbf{2.} based on workflows obtained in task 1, predict the remaining phase proportion of an ongoing period when only partial data are available; \textbf{3.} based on workflows obtained in task 1, localize procedural anomalies within new periods. Rather than directly assessing the workflow itself, we incorporate workflow flexibility by evaluating its effectiveness through downstream tasks 2 and 3. All three tasks are suitable for real-time livestream applications, with only task 1 requiring an initial observation period to infer the workflow structure.

\noindent\textbf{Evaluation Metrics.} 
For task 1, we adopt Mean Absolute Percentage Error (MAPE) to evaluate period counting accuracy. This approach aligns with previous evaluation protocols in S-PAD~\cite{dwibedi2020counting, sinha2024every, luo2024rethink, lu2024fca, li2024repetitive}, explicitly using its correct name. Specifically, suppose we have $N$ sequences, the estimated and ground truth (GT) period counts are denoted by $\{\hat{m}_1, \hat{m}_2, \cdots, \hat{m}_N\}$ and $\{m_1, m_2, \cdots, m_N, \forall m_i\ge 3\}$, respectively. The MAPE is defined as:

\begin{footnotesize}
    \begin{align}     
    \text{MAPE}=\frac{1}{N}\sum_{i=1}^{N}\frac{|\hat{m}_i-m_i|}{m_i}.
    \end{align}
\end{footnotesize} 

\noindent
To enhance long-term period assessment, we utilize the average score of Temporal Intersection over Union (tIoU)~\cite{alwassel2018action, ding2023temporal, wang2023temporal} to quantify boundary accuracy. Given that the number of estimated periods may differ from the GT, we apply Hungarian matching~\cite{kuhn1955hungarian} to identify the optimal alignment between predictions and GTs, maximizing the average scores in each sequence. Similar approaches have been applied in previous unsupervised SAS evaluations~\cite{sener2018unsupervised, kukleva2019unsupervised, kumar2022unsupervised, sarfraz2021temporally, xu2024temporally, li2024otas}. The average tIoU compares the start ($s$) and end ($e$) temporal points of the predicted and GT boundaries:

\begin{footnotesize}
    \begin{align}
        \text{tIoU}_{\text{period}} = \frac{1}{N} \sum_{i=1}^{N} \frac{1}{m_i}\sum_{j=1}^{m_i} 
        &\frac{\max\big(0, \min(e_{i,j}, \hat{e}_{i,j}) - \max(s_{i,j}, \hat{s}_{i,j})\big)}{\max(e_{i,j}, \hat{e}_{i,j}) - \min(s_{i,j}, \hat{s}_{ i,j})}. 
    \end{align}
\end{footnotesize}

\noindent
For task 2, we employ Mean Absolute Error (MAE) to compare the estimated remaining proportion value and the GT: 

\begin{footnotesize}
    \begin{align}
        \text{MAE} =\frac{1}{N}\sum_{i=1}^{N}|\hat{z_i}-z_i|,
    \end{align}
\end{footnotesize}
where $\hat{z_i}$ and $z_i$ are the estimated and GT remaining phase proportions, respectively. For task 3, we currently only assign one anomalous section per test period. Thus, unlike task 1, we do not apply Hungarian matching but instead, rely on tIoU to evaluate the overlap between predicted anomalous regions and GT in each sequence~\cite{ghoddoosian2023weakly}, as follows:

\begin{footnotesize}
    \begin{align}
        \text{tIoU}_{\text{anomaly}} &= \frac{1}{N} \sum_{i=1}^{N} 
        \frac{\max\big(0, \min(e_{i}, \hat{e}_{i}) - \max(s_{i}, \hat{s}_{i})\big)}{\max(e_{i}, \hat{e}_{i}) - \min(s_{i}, \hat{s}_{i})}. 
    \end{align}
\end{footnotesize}

\section{Our Baseline Method}
\label{sec:method}

\begin{figure}[t!]
    \centering
    \includegraphics[width=\linewidth, height=8.7cm]{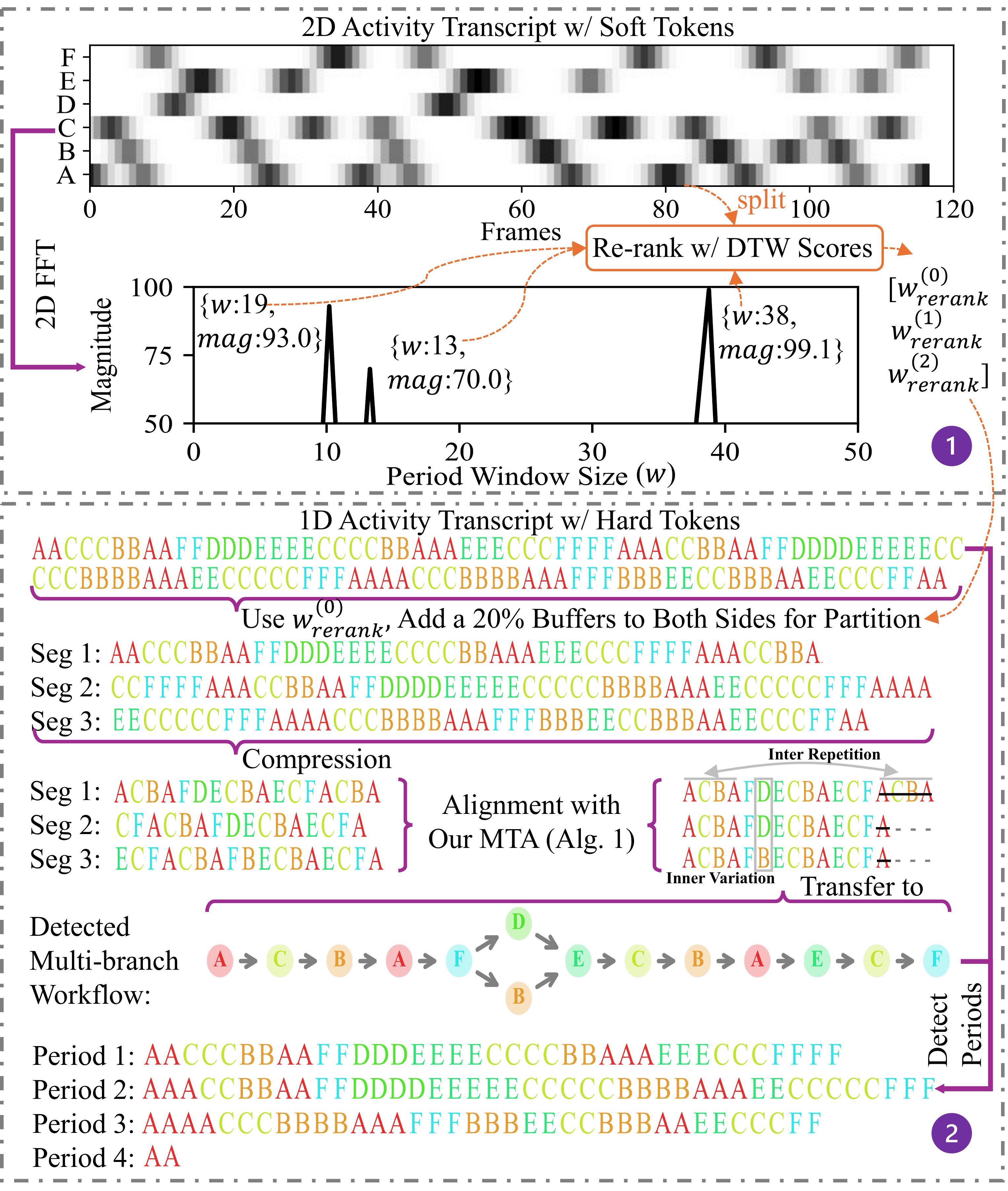}
    \caption{\textbf{Our baseline method.} In Step 1, we construct an activity transcript using soft tokens to determine the initial period window size ($w$). In Step 2, we apply our sequential mining algorithm to extract the workflow and identify the period boundaries.}
    \label{fig:baseline_method}
\end{figure}

Although human activity occurs continuously in spatiotemporal space, its features can be discretized into tokenized representations. Ideally, activity tokens might be represented as textual semantic tags~\cite{zhu2023motionbert, qu2024llms}; however, despite advances in vision foundation models~\cite{achiam2023gpt, team2023gemini, zhang2024vision, awais2025foundation}, it is challenging to fully capture nuanced spatial-temporal relationships, intricate coordination, and procedural complexities that transcend verbal descriptions~\cite{li2021towards}. For instance, text alone cannot precisely convey the dynamic motions of two hands. To effectively manage activity tokens without additional model training, an alternative approach is to represent them using a finite set of symbols~\cite{guerra2007language, li2021towards, guo2025recent}, with each symbol corresponding to a specific activity spatiotemporal state. Consequently, a transcript of human activity can be tokenized into a symbolic transcript (\ie, a string).
Given a human activity transcript of length $T$ with frame-wise spatiotemporal features $\{\mathbf{x}_t \in \mathbb{R}^{n \times 1} \}_{t=0}^{T-1}$, we perform spatiotemporal clustering to organize these features into $K$ clusters. While $K$ can be automatically determined using non-parametric clustering methods~\cite{roberts1997parametric}, we simply use K-means clustering~\cite{kodinariya2013review} for convenience. Empirically, a setting $K$ between $6$ and $14$ is sufficient to represent most workflows aimed at our task completion tracking and anomaly detection. 

After the clustering, let $\mathbf{C}_k$ be the center of the  $k$-th cluster $k = 0, 1, \ldots, K-1$. We then tokenize $\{\mathbf{x}_t\}_{t=0}^{T-1}$ into hard and soft tokens, denoted as $y^\text{hard}_t$ and $\boldsymbol{y}^\text{soft}_t$, respectively, as follows:

\vspace{-0.2cm}
\begin{footnotesize}
\begin{align}
y^\text{hard}_t &= \arg\min_{k} \|\mathbf{x}_t - \mathbf{C}_k\|;\\
\boldsymbol{y}^\text{soft}_t  &= \{d_{tk}\}_{k=0}^{K-1},
d_{tk} = \frac{\exp(-\|\mathbf{x}_t - \mathbf{C}_k\|)}{\sum_{j=0}^{K-1} \exp(-\|\mathbf{x}_t - \mathbf{C}_j\|)}.
\end{align}
\end{footnotesize}

The hard-encoded transcript $\{y^\text{hard}\}_{t=0}^{T-1}$ forms a 1D array, while the soft-encoded transcript $\{\boldsymbol{y}^\text{soft}\}_{t=0}^{T-1}$ forms a 2D matrix.
Next, we attempt to obtain a coarse period window size from our encoded activity transcript. As shown in Fig.~\ref{fig:overview},  the period cannot be detected by simply searching for peak signals within the activity transcript.  Alternatively, Fast Fourier Transform (FFT)~\cite{duhamel1990fast} and AutoCorrelation Function (ACF)~\cite{breitenbach2023method} are two prevalent approaches for detecting periods in sequential data. FFT converts a sequence into the frequency domain and predicts the period using the inverse of the frequency with the highest power. ACF assesses the similarity between a signal and its time-shifted versions, revealing repeating patterns through peaks at lag values corresponding to the period. Both FFT and ACF are proposed to work on 1D data~\cite{elfeky2005periodicity, li2010mining, breitenbach2023method}, corresponding to our $\{y^\text{hard}\}_{t=0}^{T-1}$. 
However, in highly overlapping spatiotemporal trajectories, boundary ambiguities inherited in $\{y^\text{hard}\}_{t=0}^{T-1}$ can introduce noise and impede period estimation. To reduce ambiguities, we use $\{\boldsymbol{y}^\text{soft}\}_{t=0}^{T-1}$ to incorporate the global spatiotemporal context, represented by $K$ features at each time step $t$. Next, we accommodate 2D FFT with a context marginalization to estimate the period window size $w$ in the temporal domain, which is formulated as follows:

\vspace{-0.2cm}
\begin{footnotesize}
\begin{align}
\mathcal{F}(u,v) = \sum_{t=0}^{T-1} \sum_{k=0}^{K-1} \boldsymbol{y}^\text{soft}_{k,t} \cdot \boldsymbol{e}^{-j2\pi \left( \frac{u k}{K}+\frac{v t}{T} \right)},
\end{align}
\end{footnotesize}

\noindent
where \( (u,v) \) are the frequency indices. Let \(|\cdot| \) represent the magnitude spectrum, the frequency components along the temporal axis and the corresponding magnitude are:

\vspace{-0.2cm}
\begin{footnotesize}
\begin{align}
f(v) &= \frac{v}{T}, \quad v = 0,1,\dots,T-1;\\
Mag\left(f\left(v\right)\right) &= \sum_{u=0}^{K-1} |\mathcal{F}(u, v)|.
\end{align}
\end{footnotesize}

\noindent
Since the contextual periodic signal is present in $K$ dimensions, marginalization amplifies the periodic component in $Mag\left(f\left(v\right)\right)$. We then select the top-3 dominant frequencies $\left\{f_{v,\max}^{(0)}, f_{v,\max}^{(1)}, f_{v,\max}^{(2)}\right\}$ and their corresponding period window sizes $\left\{w_{\max}^{(0)}, w_{\max}^{(1)}, w_{\max}^{(2)}\right\}$ by using:

\vspace{-0.2cm}
\begin{footnotesize}
\begin{align}
[f_{v,\max}^{(0)}, f_{v,\max}^{(1)}, f_{v,\max}^{(2)}] &= \operatorname{argsort}_{f(v)}\Big( \operatorname{Mag}\big(f(v)\big) \Big)[-3:],\\
\nonumber
\text{where} \quad w_{\max}^{(j)} &= \frac{1}{f_{v,\max}^{(j)}}, \quad f_{v,\max}^{(j)} \neq 0, \quad  j = 0,1,2.
\end{align}
\end{footnotesize}
\noindent The obtained window sizes could be further refined. We first use the window candidates to partition $\{\boldsymbol{y}^\text{soft}\}_{t=0}^{T-1}$ into segments, denoted as $\left\{\{\boldsymbol{y}^\text{soft}\}_{t=0}^{w_{\max}^{(j)}}, \{\boldsymbol{y}^\text{soft}\}_{w_{\max}^{(j)}}^{2w_{\max}^{(j)}},... \right\}$. Next, we compute the Dynamic Time Warping (DTW)~\cite{itakura2003minimum,salvador2007toward} distances between these segments to rerank the period window sizes, yielding $\left\{w_{rerank}^{(0)}, w_{rerank}^{(1)}, w_{rerank}^{(2)}\right\}$. Finally, we select $w_{rerank}^{(0)}$ as the initial period window size.

\begin{algorithm}[tbph]
    \footnotesize
    \caption{Multiple Transcript Alignment (MTA)}
    \label{alg:MTA}
    \begin{algorithmic}[1]
    \Function{MTA}{transcripts, gap\_penalty = -1}
        \State $\mathbf{F}, \mathbf{P} \gets \textnormal{InitializeMatrix}\text{(transcripts)}$  \Comment{Alg.~\ref{alg:helper_func}}
        \State dims $\gets \mathbf{F}.\text{shape}$
        \ForAll{$\text{pos} \in \text{product}(*[\text{range}(1, d) \text{ for } d \text{ in dims}])$}
            \State max\_score, best\_moves $\gets -\infty, []$
            \ForAll{neighbor in \textnormal{GetNeighbors}(pos, dims)}  \Comment{Alg.~\ref{alg:helper_func}}
                \State chars $\gets [\text{`-'}$ if $c = p$ else $\text{transcripts}[i][c-1]$ for $i, (c, p) \text{ in enumerate(zip(pos, neighbor))}]$
                \State score $\gets \mathbf{F}[\text{neighbor}] + \textnormal{ScoreMatch}(\text{chars})$  \Comment{Alg.~\ref{alg:helper_func}}
                \If{score $>$ max\_score}
                    \State max\_score $\gets$ score, best\_moves $\gets [\text{neighbor}]$
                \ElsIf{score $=$ max\_score}
                    \State best\_moves.append(neighbor)
                \EndIf
            \EndFor
            \State $\mathbf{F}[\text{pos}], \mathbf{P}[\text{pos}]  \gets \text{max\_score}, \text{best\_moves}$
        \EndFor
        \State aligned $\gets [[]  \text{for  \_ in transcripts}]$
        \State current\_pos $\gets \text{tuple}(d-1 \text{ for } d \text{ in dims})$
        \While{$\text{any}(\text{current\_pos})$ and $\mathbf{P}[\text{current\_pos}]$}
            \State prev\_pos $\gets \mathbf{P}[\text{current\_pos}][0]$
            \ForAll{$i, (c, p) \in \text{enumerate(zip(current\_pos, prev\_pos))}$}
                \State aligned[i].\text{append}($\text{transcripts}[i][c-1]$ \ \text{if} \ $c \neq p$ \ \text{else} \ `-')
            \EndFor
            \State current\_pos $\gets \text{prev\_pos}$
        \EndWhile
        \State \Return $[(\text{`'.join(seq)})[::-1] \text{ for seq in aligned}]$
    \EndFunction
    \end{algorithmic}
\end{algorithm}
\begin{algorithm}[tbph]
    \footnotesize
    \caption{Helper Functions}
    \label{alg:helper_func}
    \begin{algorithmic}[1]
    \Function{\textnormal{InitializeMatrix}}{transcripts} \Comment{Create DP score and pointer matrix}
        \State dims $\gets [\text{len(seq)} + 1 \text{ for seq in transcripts}]$ 
        \State $\mathbf{F}, \mathbf{P} \gets \text{zeros(dims)}, \text{empty(dims)}$
        \ForAll{idx in $\text{ndindex}(*\text{dims})$} $\mathbf{P}[\text{idx}] \gets []$ \EndFor
        \ForAll{idx, dim in enumerate(dims)}   
            \State $\mathbf{F}[\text{tuple([slice(None) if i = idx else 0 for i in range(len(dims))])}]$ 
            \State \hskip\algorithmicindent  $\gets \text{linspace}(0, -\text{len(transcripts)} \times \text{dim}, \text{dim})$
        \EndFor
        \State \Return $\mathbf{F}, \mathbf{P} $
    \EndFunction
    \Function{\textnormal{GetNeighbors}}{current\_pos, dims} \Comment{List valid positions in the DP matrix}
        \State neighbors $\gets []$
        \ForAll{$i \in [0, 2^{\text{len(dims)}})$}
            \ForAll{$j, pos \in \text{enumerate(current\_pos)}$}
                \If{$i \& (1\ll j)$ \textbf{and} $pos>0$} \State neighbor.append($pos - 1$)
                \Else \hspace{0.5em} neighbor.append($pos$) \EndIf
            \EndFor
            \If{$\text{len(neighbor)} = \text{len(dims)}$} \State neighbors.append($\text{tuple(neighbor)}$) \EndIf
        \EndFor
        \State \Return neighbors[1:]   
    \EndFunction
    \Function{\textnormal{ScoreMatch}}{chars}    \Comment{Compute alignment score for given transcripts}
        \If{`-' in chars} \Return $-\text{count}([c \text{ for } c \text{ in chars if } c = \text{`-'}])$
        \Else \hspace{0.5em} \Return $\sum_{i, j} (\text{chars}[i] = \text{chars}[j]) - \text{len(chars)}$ \EndIf
    \EndFunction
    \end{algorithmic}
\end{algorithm}

Although $w_{rerank}^{(0)}$ has been obtained, the period boundaries are still unknown. Some prior works assume a uniform period scale and directly use the estimated window
to segment periods. However, we argue that period window sizes can vary. Therefore, we employ the FFT-derived scale as an initial estimate and subsequently extract the workflow to pinpoint precise period boundaries (Fig.~\ref{fig:baseline_method}).

Specifically, we segment the entire transcript into multiple sections based on the estimated time window and incorporating a $20\%$ buffer at both sides to account for temporal variations. We then adapt a modified Needleman–Wunsch (NW) algorithm~\cite{needleman1970general} to detect the workflow shared across all segments. The original NW algorithm optimizes sequence alignment by minimizing gaps (\ie, `-') and maximizing matches between two sequences while considering all possible alignments. We extend the NW algorithm into a Multiple Transcript Alignment (MTA) algorithm, which jointly optimizes multiple sequence alignments through dynamic programming (DP). The detailed methodology is outlined in Algs.~\ref{alg:MTA} \& \ref{alg:helper_func}. The core idea is to fill a multi-dimensional DP matrix $\mathbf{F}$, where each cell $\mathbf{F}[\text{pos}]$ represents the optimal score to align prefixes of the input sequences ending at $\text{pos}$:
\begin{equation}
\begin{footnotesize}
\begin{aligned}
\mathbf{F}[\text{pos}] &= 
\max_{\text{prev} \in \text{GetNeighbors}(\text{pos}, \text{dims})} 
\Biggl( \mathbf{F}[\text{prev}] + \text{ScoreMatch} \biggl( \\ &\Big[ 
\quad \begin{cases} 
    \text{`-'} & \text{if } \text{pos}_i = \text{prev}_i \\ 
    \text{transcripts}[i][pos_i - 1] & \text{if } \text{pos}_i \neq \text{prev}_i 
\end{cases} 
\Big]_{i} \biggl) \Biggl)
\end{aligned}
\end{footnotesize}
\end{equation}
Due to the inclusion of buffers, redundant transcripts could be added. Thus, we implement a bidirectional traversal strategy to identify the first recurring motif (inter repetition) as the definitive boundary point. Additionally, gaps (`-') occurring before the start symbol or after the end symbol are also considered as boundary points. Elements situated beyond these demarcated boundaries are systematically eliminated to improve alignment accuracy. Moreover, we model inner variation through a multi-branch workflow to capture the complexity of divergent transcriptional pathways. 

We perform period boundary detection, completion tracking, and anomaly detection by aligning input stream tokens with workflow tokens. Specifically, we initialize two pointers: one for the workflow and one for the input stream. The workflow pointer loops within the workflow and is initialized at its end, while the input stream pointer starts at the beginning of the stream. At each time step, we compare the tokens pointed to in both sequences. If the current input token matches the workflow’s start token while the workflow pointer is at its end, this indicates that a new period has begun and the previous one has ended. As the stream progresses, we advance both pointers in synchronization, enabling continuous, online processing of all three tasks.

\section{Experiments}
\label{sec:experiments}

\begin{table*}[t!]
    \begin{tabular}{ p{0.55\linewidth} p{0.43\linewidth}}
      \begin{subtable}[b]{\hsize}
      \begin{center}
        \setlength{\tabcolsep}{0.1pt}
        \scriptsize
        \begin{tabular}{lccccccc}
          \toprule
        & \multicolumn{3}{c}{Task 1 (PD)} & \multicolumn{2}{c}{Task 2 (CT)} & \multicolumn{2}{c}{Task 3 (AD)} \\
        \cmidrule(lr){2-4} \cmidrule(lr){5-6} \cmidrule(lr){7-8}
        & MAPE $\downarrow$ & $\text{tIoU}_{\text{period}}$ $\uparrow$ & Runtime $\downarrow$ & MAE $\downarrow$ & Runtime $\downarrow$ & $\text{tIoU}_{\text{anomaly}}$  $\uparrow$ & Runtime $\downarrow$\\ \midrule
        \multicolumn{8}{l}{\textcolor{gray}{Unsupervised sequential periodic detection (applicable to task 1: Period Detection (PD))}} \\
        ClaSP~\cite{ermshaus2023clasp} & 1.284 & 0.331 &  \best{0.4X} & - & - & - & -\\ 
        CFDAutoperiod~\cite{puech2020fully} & 8.396  & 0.146  & 6X & - & - & - & - \\
        1D FFT~\cite{hyndman2018forecasting} & 2.746  & 0.399 & \second{0.6X} & - & - & - & - \\
        ACF~\cite{breitenbach2023method} & 2.432 & 0.439 &  4X & - & - & - & -\\ \hline
        \multicolumn{8}{l}{\textcolor{gray}{Unsupervised sequential anomaly detection (applicable to task 3: Anomaly Detection (AD))}} \\
       
        TadGAN~\cite{geiger2020tadgan,alnegheimish2022sintel} & - & - & - & - & - & 0.066 & $>$100X \\
        AER~\cite{wong2022aer,alnegheimish2022sintel} & - & - & - & - & - & 0.068 & $>$100X \\
        AnomalTrans~\cite{xu2022anomaly} & - & - & - & - & - & 0.085 & $>$100X \\ \hline
        \multicolumn{8}{l}{\textcolor{gray}{Unsupervised sequential reasoning (applicable to all tasks: PD, Completion Tracking (CT), AD)}} \\ 
        GPT-4o~\cite{achiam2023gpt,hurst2024gpt} & 0.038 & 0.882 & $\approx$5X & 0.388 & \second{$\approx$5X} & 0.173 & \second{$\approx$5X} \\
        Gemini-2.5 Pro~\cite{comanici2025gemini} & \second{0.030}  & \second{0.915} & $\approx$10X & \second{0.286}  & $\approx$10X  & \second{0.256} & $\approx$10X  \\ \hline
        \textbf{Our Baseline}  & \best{0.025}   &  \best{0.937} & 1X & \best{0.106} & \best{1X} & \best{0.423} & \best{1X} \\ \bottomrule
        \end{tabular}
        \subcaption{Comparison of unsupervised methods on our 3 benchmark tasks. Workflows are extracted from task 1 and used for tasks 2 and 3.}
        \label{table:comparison_our_benchmark}
      \end{center}
      \end{subtable}
    &
      \begin{subtable}[b]{\hsize}
      \begin{center}
        \setlength{\tabcolsep}{0.15pt}
        \scriptsize
        \begin{tabular}{cccccc}
          \toprule
          \multirow{2}{*}{\begin{tabular}[c]{@{}c@{}} Token for $w$\\Initialization\end{tabular}} &~ \multirow{2}{*}{\begin{tabular}[c]{@{}c@{}}Buffer\\Size\end{tabular}} & \multicolumn{2}{c}{Task 1 (PD)} & \multicolumn{1}{c}{Task 2 (CT)} & \multicolumn{1}{c}{Task 3 (AD)} \\
          \cmidrule(lr){3-4} \cmidrule(lr){5-5} \cmidrule(lr){6-6}
          & & MAPE $\downarrow$ & $\text{tIoU}_{\text{period}}$ $\uparrow$   & MAE $\downarrow$  & $\text{tIoU}_{\text{anomaly}}$  $\uparrow$ \\ \midrule
          \multicolumn{6}{l}{\ours{w/ window size re-ranking}} \\
          $\{y^\text{hard}\}_{i=0}^{T-1}$ & $10\%$  &\vline \white{0.149}   &  0.842 & 0.164  & 0.361  \\ 
          $\{y^\text{hard}\}_{i=0}^{T-1}$ & $20\%$  &\vline \white{0.147}   &  0.854 & 0.160  & 0.374  \\ 
          $\{y^\text{hard}\}_{i=0}^{T-1}$ & $40\%$  &\vline  \white{0.150}   &  0.833 & 0.167  & 0.358  \\ 
          $\{\boldsymbol{y}^\text{soft}\}_{i=0}^{T-1}$ & $10\%$  &\vline  \second{0.026}   &  \second{0.930} & \second{0.108}  & \second{0.422}  \\ 
          \ours{$\{\boldsymbol{y}^\text{soft}\}_{i=0}^{T-1}$} & \ours{$20\%$}  &\vline  \best{0.025}   &  \best{0.937} & \best{0.106}  & \best{0.423}  \\ 
          $\{\boldsymbol{y}^\text{soft}\}_{i=0}^{T-1}$ & $40\%$  &\vline  \white{0.036}   &  0.928 & 0.124  & 0.401  \\ 
          \hline
          \multicolumn{6}{l}{\textcolor{gray}{w/o window size re-ranking}} \\
          $\{y^\text{hard}\}_{i=0}^{T-1}$ & $20\%$  &\vline \white{1.392}   &  0.375 & 0.623  & 0.152  \\ 
          $\{\boldsymbol{y}^\text{soft}\}_{i=0}^{T-1}$ & $20\%$  &\vline \white{0.932}   &  0.539 & 0.342  & 0.207  \\ 
          \bottomrule
        \end{tabular}
        \subcaption{Ablation studies on our baseline (re-ranking $\&$ token $\&$ buffer). Our standard baseline parameters are rendered in \ours{green}.}
        \label{table:ablation_studies}
      \end{center}
      \end{subtable}
    \end{tabular}
    \vspace{-0.4cm}
    \caption{\textbf{Evaluations on our benchmark.} We apply a constant $K=10$ for tokenization and use the identical transcripts for all methods as inputs. Runtime on each task of our benchmark is normalized to our baseline (1X). }
    \label{tab:our_benchmark_evaluation}
    \vspace{-0.2cm}
  \end{table*}

\begin{figure*}[th!]
  \centering
  \includegraphics[width=\linewidth, height=2.5cm]{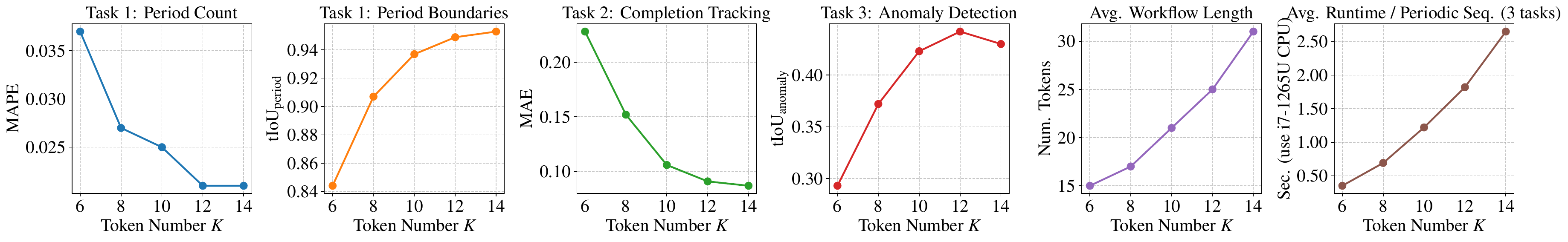}
  \vspace{-0.75cm}
  \caption{\textbf{Ablation studies with various $K$.} Larger values improve performance at the expense of higher computational cost.}
  \label{fig:ablation_study_plots_K}
  \vspace{-0.4cm}
\end{figure*}

\begin{figure*}[th!]
  \centering
  \includegraphics[width=\linewidth, height=3.6cm]{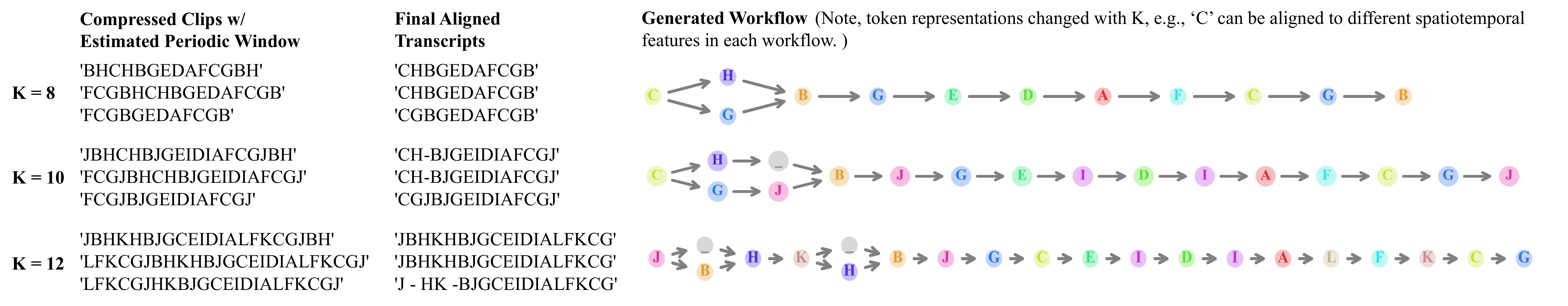}
  \vspace{-0.7cm}
  \caption{\textbf{An example of workflows generated for the same activity sequence with different values of $K$.} The symbol `\textcolor{gray}{\textunderscore}' represents a skipped token. As $K$ increases, the workflows become more detailed and extended.}
  \label{fig:workflow_results_ablation_K}
  \vspace{-0.5cm}
\end{figure*}

\begin{figure*}[ht!]
\centering
\begin{minipage}[b]{0.48\textwidth}
    \centering
    \includegraphics[width=\textwidth, height=6.3cm]{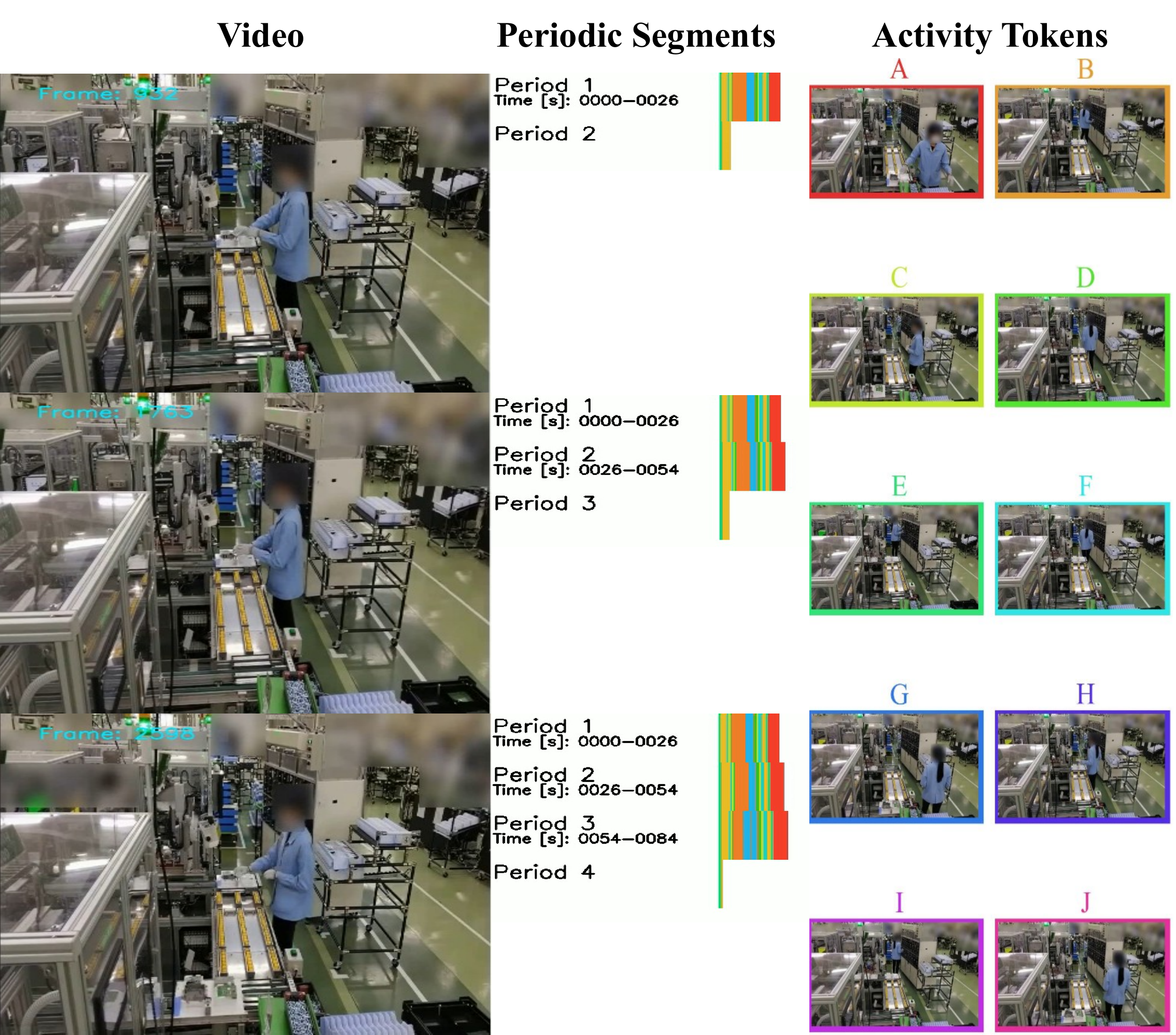}
    \vspace{-0.55cm}
    \caption{\textbf{Illustration of applying our baseline to analyze production workflows} (see supp. video for details).}
    \vspace{-0.35cm}
    \label{fig:app_demo}
\end{minipage}
\hfill
\begin{minipage}[b]{0.49\textwidth}
    \centering
    \includegraphics[width=\textwidth, height=6.3cm]{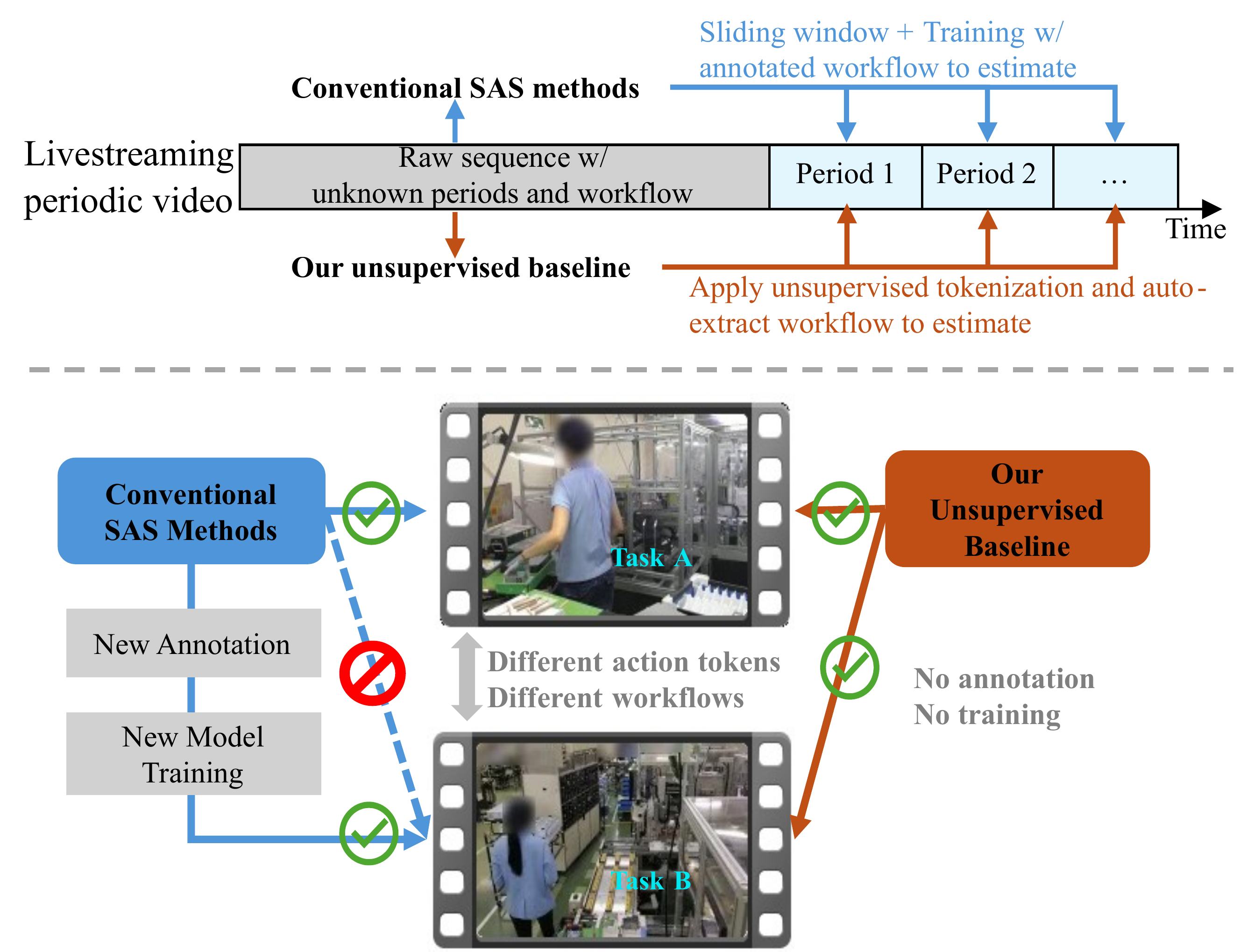}
    \vspace{-0.55cm}
    \caption{\textbf{Comparison of deployment costs.} \textbf{Top:} Deployment for one livestream. \textbf{Bottom:} Deployment for two different livestreams. }
    \vspace{-0.45cm}
    \label{fig:app_cost}
\end{minipage}
\end{figure*}

We conducted a series of experiments using the metrics introduced in Sec.~\ref{sec:dataset_and_metrics} and the activity tokenization method introduced in Sec.~\ref{sec:method}. For task 1, we follow the common unsupervised evaluation protocols and apply all samples as the test set, allowing the model to learn the underlying patterns and perform estimations without any training data. The workflows extracted from task 1 are then used for tasks 2 and 3. We evaluate our baseline against traditional unsupervised mining methods, supervised PAD and SAS approaches, and LLMs.
Our results are presented in Tabs.~\ref{tab:our_benchmark_evaluation},~\ref{tab:comparison_short_term}, and~\ref{tab:comparison_PAD_on_our_benchmark}, and Figs.~\ref{fig:ablation_study_plots_K},~\ref{fig:workflow_results_ablation_K},~\ref{fig:app_demo}, and~\ref{fig:app_cost}. The best and second-best results are marked in \best{purple} and \second{orange}, respectively. Below, we summarize our key findings from the results.

\noindent\textbf{Our tokenization enables LLM-based periodic analysis.} 
Given our tokenized transcripts (Fig.~\ref{fig:baseline_method}), we apply both GPT-4o\cite{achiam2023gpt, hurst2024gpt} and Gemini-2.5~\cite{comanici2025gemini} to analyze our benchmark, using the task descriptions directly as prompts (Tab.~\ref{table:comparison_our_benchmark}). While they are comparable to our baseline method on task 1, they struggle with tasks 2 and 3. One possible explanation is that when provided with hundreds of repetitive, low-contrast tokens, LLMs can identify the repetitive patterns but may have difficulty extracting the long-term workflow used for tasks 2 and 3. 

\noindent\textbf{Traditional unsupervised sequential mining methods struggle on our benchmark.} Unlike previous datasets used for periodicity and anomaly detection studies, our dataset features low-contrast periodic patterns and complex workflows that significantly challenge existing approaches (Tab.~\ref{table:comparison_our_benchmark}). For instance, in period counting, the MAPE scores of unsupervised sequential periodic detection methods we evaluated~\cite{ermshaus2023clasp,puech2020fully,hyndman2018forecasting,breitenbach2023method} are over $50$ times higher than that of our baseline method. Similarly, in anomaly detection, the representative methods we assessed~\cite{geiger2020tadgan,wong2022aer,xu2022anomaly} perform nearly at the level of random guessing. Therefore, our benchmark presents a novel challenge to the research community, encouraging the development of new solutions.

\noindent\textbf{Ablation studies on baseline parameters.} We demonstrate the effectiveness of employing our initialized window re-ranking, soft tokenization, and $20\%$ buffer size for period window estimation, as detailed in Tab.~\ref{table:ablation_studies}. We also explore the impact of varying the parameter $K$ in Fig.~\ref{fig:ablation_study_plots_K}. Changes in $K$ influence the representation of the extracted workflows. In complicated scenarios, a larger $K$ leads to fine-grained tokens and may remove ambiguities with a smaller $K$, thereby improving performance on our benchmark. However, this improvement comes with increased computational cost. 
Fig.~\ref{fig:workflow_results_ablation_K} illustrates the extracted workflows for various $K$ values. As $K$ increases, the workflows become more detailed and extended. Notably, the extracted workflows exhibit multi-branch structures that effectively capture diverse approaches within the activity while accounting for spatiotemporal variations and noise.

\begin{wraptable}{r}{0.23\textwidth} 
    \hspace{-4.5\baselineskip}
    \centering
    \setlength{\tabcolsep}{0.1pt}
    \scriptsize
    \begin{tabular}{lc}
        \toprule
    & $\text{tIoU}_{\text{anomaly}}$(Task 3)$\uparrow$ \\ 
        \hline
        \multicolumn{2}{l}{\textcolor{gray}{Supervised (w/ task 1 labeled data)}} \\
        ASFormer~\cite{asformer2021}  & 0.294 \\
        FACT~\cite{lu2024fact}  & \second{0.312}\\
        \midrule
        \multicolumn{2}{l}{\textcolor{gray}{Unsupervised (w/ task 1 unlabeled data)}} \\
        HVQ~\cite{spurio2024hierarchical}  & 0.087\\
        \textbf{Our Baseline}   &   \best{0.423}   \\ 
        \bottomrule
    \end{tabular}
    \vspace{-0.2cm}
    \hspace{-4.5\baselineskip}
        \caption{Comparison of SAS methods for task 3.}
    \label{tab:comparison_SAS_on_our_benchmark}
    \vspace{-0.4cm}
\end{wraptable}
\noindent\textbf{Our baseline outperforms supervised SAS methods on task 3.} 
Although existing SAS methods cannot handle periodic workflow detection in task 1, they can be adapted to task 3 by training on our annotated task 1 data to estimate workflows and detect anomalies. We compare our baseline with both supervised and unsupervised SAS methods in Tab.~\ref{tab:comparison_SAS_on_our_benchmark}. Our baseline outperforms supervised methods. The unsupervised SAS method HVQ~\cite{spurio2024hierarchical}, which extracts patterns from unlabeled periodic data without explicitly modeling periodicity, fails to generalize to task 3. As the SAS benchmarks do not involve periodicity, which is beyond the scope of our study, we do not evaluate our baseline on them.

\noindent\textbf{Our baseline outperforms on the S-PAD benchmark, but supervised S-PAD methods underperform on our benchmark.} 
Tab.~\ref{tab:comparison_short_term} illustrates that our baseline method, enhanced with advanced activity tokenization and sequential modeling, effectively detects periodic patterns in an unsupervised manner. This approach surpasses several supervised methods~\cite{hu2022transrac, dwibedi2024short, yao2023poserac, sinha2024every} on the traditional S-PAD benchmark RepCount~\cite{hu2022transrac}. In contrast, representative conventional S-PAD methods~\cite{yao2023poserac, sinha2024every} perform poorly on our benchmark---even when provided with supervision---highlighting their limitations in modeling long-term periodic patterns. These results demonstrate that our method is capable of effectively handling both short- and long-term periodic structures.
\begin{table}[h!]
  \begin{minipage}[b]{0.42\linewidth}
    \centering
    \setlength{\tabcolsep}{0.1pt}
    \scriptsize
    \begin{tabular}{lc}
      \toprule
      & MAPE (Task 1) $\downarrow$  \\ \hline
      \multicolumn{2}{l}{\textcolor{gray}{Supervised}} \\
      TransRAC~\cite{hu2022transrac} & 0.443  \\
      RepNet~\cite{dwibedi2024short} & 0.331  \\
      SkimFocus~\cite{zhao2024skim} & 0.249  \\
      PoseRAC~\cite{yao2023poserac} & 0.236  \\
      ESCounts~\cite{sinha2024every} & \second{0.213} \\
      \midrule
      \multicolumn{2}{l}{\textcolor{gray}{Unsupervised}} \\
      \textbf{Our Baseline}   &   \best{0.156}   \\ \bottomrule
    \end{tabular}
    \caption{Our baseline on the S-PAD benchmark. }
    \label{tab:comparison_short_term}
    \vspace{-0.5cm} 
  \end{minipage}\hfill 
  \begin{minipage}[b]{0.56\linewidth}
    \centering
    \setlength{\tabcolsep}{1pt} 
    \scriptsize
    \begin{tabular}{lc}
      \toprule
      & MAPE (Task 1) $\downarrow$  \\ \hline
      \multicolumn{2}{l}{\textcolor{gray}{Supervised}} \\
      PoseRAC~\cite{yao2023poserac} & 0.738\\
      ESCounts~\cite{sinha2024every} & \second{0.673} \\
      \midrule
      \multicolumn{2}{l}{\textcolor{gray}{Unsupervised}} \\
      \textbf{Our Baseline}   &   \best{0.023}   \\ \bottomrule
    \end{tabular}
    \vspace{-0.2cm} 
    \caption{S-PAD methods on our benchmark. 380 samples with video inputs are used for task 1 evaluation. We use 4-fold cross-validation for supervised methods.}
    \label{tab:comparison_PAD_on_our_benchmark}
    \vspace{-0.5cm} 
  \end{minipage}
\end{table}

\noindent\textbf{Our baseline is efficient in real-world deployments.} We validate the effectiveness of our approach by deploying it on a factory production line (see Fig.~\ref{fig:app_demo}). Our method accurately identifies the production workflow patterns from raw periodic video streams, providing actionable insights for tracking the remaining phase proportion of an ongoing cycle and detecting procedural anomalies.
We also conceptually analyze the deployment costs of our method relative to supervised alternatives (see Fig.~\ref{fig:app_cost}). Overall, our method supports seamless deployment across various production lines, even as procedures and schedules change frequently, demonstrating both its scalability and efficiency in real-world applications.

\section{Conclusion}
We introduce a pioneering benchmark and baseline for the unsupervised detection of long-term periodic spatiotemporal workflows in human activities. 
We show its practical applications through three evaluation tasks: unsupervised period detection, task completion tracking, and anomaly detection. Our baseline, enhanced with advanced activity tokenization and sequential modeling, effectively detects periodic patterns in an unsupervised manner. We believe our work lays the foundation for future research and applications in long-term periodic human activity analysis.

{
    \small
    \bibliographystyle{ieeenat_fullname}
    \bibliography{main}
}

\end{document}